\documentclass{article}
\usepackage{graphicx}
\usepackage{subfigure}
\usepackage{url}
\usepackage{float}
\usepackage{multirow}
\usepackage{array}
\usepackage{hyperref}

\makeatletter
    \newcommand\figcaption{\def\@captype{figure}\caption}
    \newcommand\tabcaption{\def\@captype{table}\caption}
\makeatother

\begin{document}

\title{A Classification of 6R Manipulators}
\author{Ming-Zhe Chen 
\\Qwest Communications Inc.
\\ 5751 Sells Mill Dr 
        \\ Dublin, OH 43017, USA}
\maketitle

\begin{abstract}
This paper presents a classification of generic 6-revolute jointed (6R)
manipulators using homotopy class of their critical point manifold. 
A part of classification is listed in this paper because of the 
complexity of homotopy class of 4-torus.
The results of this classification will serve future research of 
the classification and topological properties of maniplators joint space 
and workspace. 
\end{abstract}

\section{Introduction}
In robotics, the topological properties of manipulator joint space and workspace are related
to the singularities. A lot of literature deal with the manipulator singularities 
\cite{burdick} \cite{wenger} \cite{gupta} \cite{kumar} \cite{karger} \cite{sugimoto}
\cite{litvin} \cite{krzysztof} \cite{chen} \cite{kieffer} \cite{selig} and references therein.
The kinematics of manipulator can be described as a smooth map:
\[ \Phi : X \mapsto Y \]
where X denotes the space of joint variables ($x_1$ $x_2$ $x_3$ $x_4$ $x_5$ $x_6$) for 
6R manipulator and Y is the workspace 
of positions and orientations of the end-effector.
Manipulator with m revolte joint, the joint space is
\[ X = T^m \]
where $T^m$ denotes the m-dimensional torus.
The workspace Y can be special euclidean group SE(3):
\[ SE(3) \equiv SO(3) \times R^3 \]
which is a Lie group, that is a semidirect product of the special orthogonal group
SO(3) and $R^3$.
From the map $\Phi$ we can work out the Jacobian, which is the linear map k 
on the tangent spaces:
\[ k: R^m \mapsto R^n \]
where m is the dimension of joint space and n is the dimension of workspace,
in this paper m = n = 6.
The Jacobian J will be related to:
\[ \left( \begin{array}{cc} \omega \\ v \end{array} \right) = J\dot{X} \]
where $\omega$ is the angular velocity of the end-effector, v is the velocity of 
the end-effector and $\dot{X}$ is the joint rate.
Let r = rank(J), then the singular set for the map k can be defined as:
\[ S_k = \{ x \in R^m |\; r < min(m, n) \} \]
The singular set can also be decomposed into strata \cite{golubitsky} \cite{krzysztof}:
\[ S_k = \bigcup_{c=1}^{min(m,n)} S_c \]
where 
\[ S_c = \{x \in R^m |\; r = min(m, n) - c \} \]
where the number c is called the corank of the map k.
A manipulator is said generic if its singularities are generic, that is, 
if they form smooth manifolds in joint space.
Non-generic manipulaters often arise from geometric simplication condition in the structue.
Generic manipulators have only ordinary singularities \cite{burdick} \cite{wenger} \cite{karger}, 
that is, the singular set (manifolds) behaves like a regular surface, i.e., corank c = 1.
In this paper, we deal with generic 6R manipulators. So the higher order singularities 
do not exist for this kind of 6R manipulators, which can be interpreted as bifurcation. 
In fact, all possible singular configurations are the zero set of det(J):
\[ det(J) = 0 \]
After some algebraic manipulation, letting $q_1 = tan(x_1/2)$, $q_2 = tan(x_2/2)$, 
$q_3 = tan(x_3/2)$, $q_4 = tan(x_4/2)$, $q_5 = tan(x_5/2)$, $q_6 = tan(x_6/2)$,  
above equation can be converted to the form:
\[ f_1(q_2,q_3,q_4)q_5^8 + \dots = 0 \]
or
\[ f_2(q_2,q_3,q_5)q_4^{10} + \dots = 0 \]
or
\[ f_3(q_2,q_4,q_5)q_3^{12} + \dots = 0 \]
or
\[ f_4(q_3,q_4,q_5)q_2^{14} + \dots = 0 \]
The result is too complicated to be displayed here. We just show the first term of 
the equation.  In above equations, we use Denavit-Hartenberg parameters.
The above equations may have "zero at infinity" (singularities at infinity).
We do not pay extra attention to it since these singularities are a part of 
the generic singular surface.
Since above equations are independent of $q_1$ and $q_6$, the singular surface is
projected onto torus $T^4$. The singular surface forms branches, i.e., the connected 
components, on the surface of $T^4$. In the following section we will classify the 
manipulator by the singular surface forming branches, which are the group of
homotopy class in $T^4$.

\section{A Classification of 6R Manipulators}
Singular surface can be characterized by their fundemental group
of homotopy class in $T^4$. The fundemental group of $T^4$ is the group of loop 
equivalence classes, denoted by $\pi_1(T^4)$ \cite{hatcher}:
\[ \pi_1(T^4) = \pi_1(T^2) \times \pi_1(T^2) = Z \times Z \times Z \times Z \]
where Z is the set of integer.
Each element of $\pi_1(T^4)$, which is a set of homotopicly equivalent singular
surfaces, can be labeled by integers ($I_2$,$I_3$,$I_4$,$I_5$), which characterize
how many integral times the "curve" "wrap around" the generator of $T^4$.
It is known a class of 6R manipulators which have the same homotopy class will
have similar topological properties in the their joint spaces.
Singular manifolds can divide the joint space of 6R manipulator in at least two
singularity-free domain called c-sheets.  A single singular manifold branch, 
which can cut the joint space into two c-sheets, is said to be separating;
otherwise, it is said non-separating, which must combine with other branch to 
divide the joint space.
In the following, we enumerate the branch homotopy classes.
\begin{flushleft}
(1) (0,0,0,0) homotopy class: the branch is separating and can appear alone.
The result is due to the topology of the torus.
\end{flushleft}
\begin{flushleft}
(2) 2(0,0,0,0), 3(0,0,0,0) and 4(0,0,0,0) homotopy classes: the branches are
 separating.
\end{flushleft}
\begin{flushleft}
But more than 4 coexisting (0,0,0,0) branches would yield more than 8 intersections
with the generator, here 8 is the minimum of the maximum allowable times 
around the generator.   
(1,0,0,0) or (0,1,0,0) or (0,0,1,0) or (0,0,0,1) cannot divide the torus, so
they cannot appear alone.
($I_2$,0,0,0) or (0,$I_3$,0,0) or (0,0,$I_4$,0) or (0,0,0,$I_5$), 
when $I_2 > 1$ or $I_3 > 1$ or $I_4 > 1$ or $I_5 > 1$, are impossible
because these helical curves must go backward, which cannot be done without
self intersections. 
Since $\pi_1(T^4) = \pi_1(T^2) \times \pi_1(T^2)$, we can consider the 
singular surfaces have two branches each of which is on one torus $T^2$.
Possible two branches are ($I_2$,$I_3$,0,0) + (0,0,$I_4$,$I_5$), here
$I_2 \leq 12$, $I_3 \leq 14$, $I_4 \leq 8$, and $I_5 \leq 10$, because 
the degrees of $q_2$, $q_3$, $q_4$, and $q_5$ in the equation det(J) = 0
set up the maximum allowable times for $I_2$, $I_3$, $I_4$, and $I_5$ 
around the generator of $T^2$. 
But either ($I_2$,$I_3$,0,0) or (0,0,$I_4$,$I_5$) cannot be separating
because they form helical closed bands.  
In paper \cite{wenger}, there are eight homotopy classes for torus $T^2$, 
that is, (0,0), 2(0,0), (0,0)+2(1,0), 2(1,0), 4(1,0), 2(0,1), 2(1,1), and 2(2,1).
We denote these classes to be H2. So following classes are homotopy classes
which are separating branches:
\end{flushleft}
\begin{flushleft}
(3) (H2,0,0) + (0,0,$I_4$,$I_5$), ($I_2$,$I_3$,0,0) + (0,0,H2) and
(H2,0,0) + (0,0,H2) homotopy classes: the branches are separating. 
\end{flushleft}
\begin{flushleft} 
For the torus $T^2$ mentioned in this paper, there are more separating homotopy classes.
For torus ($I_2$,$I_3$,0,0), (11,14,0,0) + (1,0,0,0), (10,14,0,0) + 2(1,0,0,0), 
(9,14,0,0) + 3(1,0,0,0), (8,14,0,0) + 4(1,0,0,0), (7,14,0,0) + 5(1,0,0,0), 
(6,14,0,0) + 6(1,0,0,0), (5,14,0,0) + 7(1,0,0,0), (4,14,0,0) + 8(1,0,0,0), 
(3,14,0,0) + 9(1,0,0,0), (2,14,0,0) + 10(1,0,0,0), (1,14,0,0) + 11(1,0,0,0) 
and (12,13,0,0) + (0,1,0,0), (12,12,0,0) + 2(0,1,0,0), (12,11,0,0) + 3(0,1,0,0), 
(12,10,0,0) + 4(0,1,0,0), (12,9,0,0) + 5(0,1,0,0), (12,8,0,0) + 6(0,1,0,0), 
(12,7,0,0) + 7(0,1,0,0), (12,6,0,0) + 8(0,1,0,0), (12,5,0,0) + 9(0,1,0,0), 
(12,4,0,0) + 10(0,1,0,0), (12,3,0,0) + 11(0,1,0,0), (12,2,0,0) + 12(0,1,0,0), 
(12,1,0,0) + 13(0,1,0,0).
are separating homotopy classes.
We denote these classes to be H3. 
Similarly, for torus (0,0,$I_4$,$I_5$), (0,0,7,10) + (0,0,1,0), (0,0,6,10) + 2(0,0,1,0), 
(0,0,5,10) + 3(0,0,1,0), (0,0,4,10) + 4(0,0,1,0), (0,0,3,10) + 5(0,0,1,0), 
(0,0,2,10) + 6(0,0,1,0), (0,0,1,10) + 7(0,0,1,0) and (0,0,8,9) + (0,0,0,1), 
(0,0,8,8) + 2(0,0,0,1), (0,0,8,7) + 3(0,0,0,1), (0,0,8,6) + 4(0,0,0,1), 
(0,0,8,5) + 5(0,0,0,1), (0,0,8,4) + 6(0,0,0,1), (0,0,8,3) + 7(0,0,0,1), 
(0,0,8,2) + 8(0,0,0,1), (0,0,8,1) + 9(0,0,0,1).
are separating homotopy classes.
We denote these classes to be H4.  So,
\end{flushleft}
\begin{flushleft}
(4) H3 and H4 homotopy classes: the branches are separating. 
\end{flushleft}
\begin{flushleft}
There are other combinations such as (0,0,6,8) + (0,0,1,0) etc. we cannot enumerate here. 
But the combination like (0,0,7,9) + (0,0,1,0) + (0,0,0,1) would lead to 
intersecting branches: (0,0,1,0) and (0,0,0,1), and it cannot exist. 
In fact, in order to get a class of 6R manipulators which have the same homotopy class,
i.e., to get a set of parameters of the manipulators, we must solve 
the equation det(J) = 0 under some conditions.  
For example, for (0,0,0,0) homotopy class, we must solve above equation
to get a set of parameters under the conditions:
\[ -\pi < q_2 < \pi \]
\[ -\pi < q_3 < \pi \]
\[ -\pi < q_4 < \pi \]
\[ -\pi < q_5 < \pi \]
and for 2(0,0,0,0) homotopy class, we must solve equation det(J) = 0 
to get two branches solution and a set of parameters under above conditions.
There are still a lot of combinations we cannot enumerate here. It indicates the 
complexity of the joint space of 6R manipulator.  
Recently, there are some papers \cite{farber1} \cite{farber2} \cite{altafini}
\cite{aronov} dealing with the topological properties of 
manipulators, but we still have little understanding of the topological properties 
of joint space and workspace of the manipulator.
\end{flushleft}
 
\bibliography{mzchen10}
\bibliographystyle{plain}
\end{document}